\documentclass[10pt,twocolumn,letterpaper]{article}

\usepackage{iccv}
\usepackage{times}
\usepackage{epsfig}
\usepackage{graphicx}
\usepackage{amsmath}
\usepackage{amssymb}
\usepackage{comment}
\usepackage{caption}
\usepackage{subcaption}
\usepackage[sort]{cite}
\usepackage{booktabs}

\definecolor{tabutter}{rgb}{0.98824, 0.91373, 0.30980}		
\definecolor{ta2butter}{rgb}{0.92941, 0.83137, 0}		
\definecolor{ta3butter}{rgb}{0.76863, 0.62745, 0}		

\definecolor{taorange}{rgb}{0.98824, 0.68627, 0.24314}		
\definecolor{ta2orange}{rgb}{0.96078, 0.47451, 0}		
\definecolor{ta3orange}{rgb}{0.80784, 0.36078, 0}		

\definecolor{tachocolate}{rgb}{0.91373, 0.72549, 0.43137}	
\definecolor{ta2chocolate}{rgb}{0.75686, 0.49020, 0.066667}	
\definecolor{ta3chocolate}{rgb}{0.56078, 0.34902, 0.0078431}	

\definecolor{tachameleon}{rgb}{0.54118, 0.88627, 0.20392}	
\definecolor{ta2chameleon}{rgb}{0.45098, 0.82353, 0.086275}	
\definecolor{ta3chameleon}{rgb}{0.30588, 0.60392, 0.023529}	

\definecolor{taskyblue}{rgb}{0.44706, 0.56078, 0.81176}		
\definecolor{ta2skyblue}{rgb}{0.20392, 0.39608, 0.64314}	
\definecolor{ta3skyblue}{rgb}{0.12549, 0.29020, 0.52941}	

\definecolor{taplum}{rgb}{0.67843, 0.49804, 0.65882}		
\definecolor{ta2plum}{rgb}{0.45882, 0.31373, 0.48235}		
\definecolor{ta3plum}{rgb}{0.36078, 0.20784, 0.4}		

\definecolor{tascarletred}{rgb}{0.93725, 0.16078, 0.16078}	
\definecolor{ta2scarletred}{rgb}{0.8, 0, 0}			
\definecolor{ta3scarletred}{rgb}{0.64314, 0, 0}			

\definecolor{taaluminium}{rgb}{0.93333, 0.93333, 0.92549}	
\definecolor{ta2aluminium}{rgb}{0.82745, 0.84314, 0.81176}	
\definecolor{ta3aluminium}{rgb}{0.72941, 0.74118, 0.71373}	

\definecolor{tagray}{rgb}{0.53333, 0.54118, 0.52157}		
\definecolor{ta2gray}{rgb}{0.33333, 0.34118, 0.32549}		
\definecolor{ta3gray}{rgb}{0.18039, 0.20392, 0.21176}		

\newcommand{\shortcite}[1]{\cite{#1}}
\newcommand{\ignorethis}[1]{}
\usepackage[pagebackref=true,breaklinks=true,letterpaper=true,colorlinks,bookmarks=false]{hyperref}
\iccvfinalcopy 

\def\iccvPaperID{0370} 

\ificcvfinal\fi

\begin{document}

\title{Learning a Discriminative Model for the Perception of Realism \\
in Composite Images}

\author{Jun-Yan Zhu\\
UC Berkeley\\
\and
Philipp Kr\"ahenb\"uhl\\
UC Berkeley\\
\and
Eli Shechtman\\
Adobe Research\\
\and
Alexei A. Efros\\
UC Berkeley\\
}

\maketitle

\begin{abstract}

What makes an image appear realistic?
In this work, we are looking at this question from a data-driven perspective, by learning the perception of visual realism directly from large amounts of unlabeled data.
In particular, we train a Convolutional Neural Network (CNN) model that distinguishes natural photographs from automatically generated composite images.
The model learns to predict visual realism of a scene in terms of color, lighting and texture compatibility, without any human annotations pertaining to it.
Our model outperforms previous works that rely on hand-crafted heuristics for the task of classifying realistic vs. unrealistic photos.
Furthermore, we apply our learned model to compute optimal parameters of a compositing method, to maximize the visual realism score predicted by our CNN model. We demonstrate its advantage against existing methods via a human perception study.

\end{abstract}

\section{Introduction}
\label{intro}

The human ability to very quickly decide whether a given image is ``realistic'', {\em i.e.} a likely sample from our visual world, is very impressive.
Indeed, this is what makes good computer graphics and photographic editing so difficult.
So many things must be ``just right" for a human to perceive an image as realistic, while a single thing going wrong will likely hurtle the image down into the Uncanny Valley~\cite{mori2012uncanny}.

Computers, on the other hand, find distinguishing between ``realistic'' and ``artificial'' images incredibly hard.
Much heated online discussion was generated by recent results suggesting that image classifiers based on Convolutional Neural Network (CNN) are easily fooled by random noise images~\cite{szegedy2013intriguing,nguyen2014deep}. But in truth, no existing method (deep or not) has been shown to reliably tell whether a given image resides on the manifold of natural images. This is because the spectrum of unrealistic images is much larger than the spectrum of natural ones.  Indeed, if this was not the case, photo-realistic computer graphics would have been solved long ago.

\begin{figure}[tr]
  \begin{center}
 \includegraphics[width=1.0\linewidth]{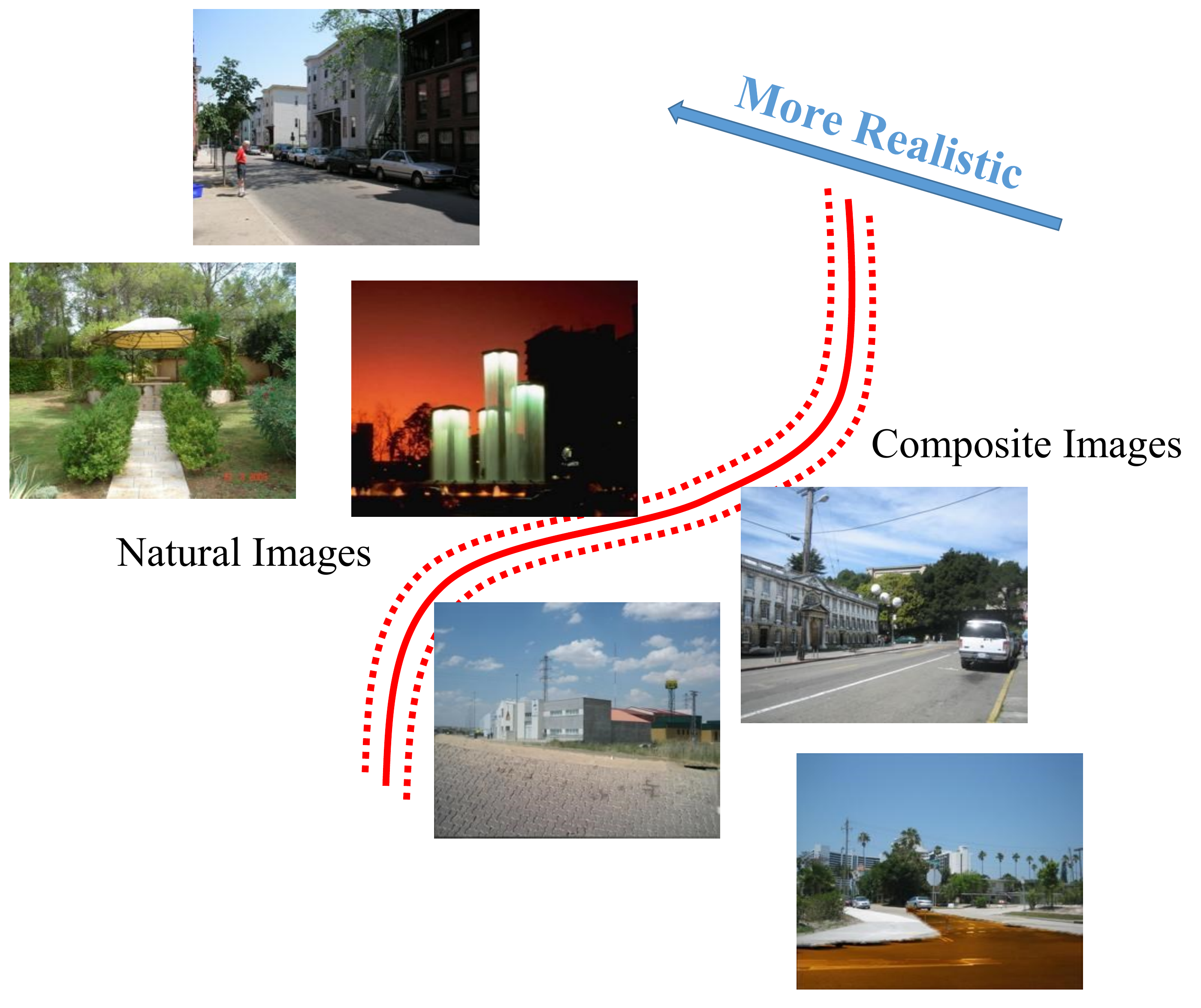}
  \end{center}
 \vspace{-5 mm}
  \caption{We train a discriminative model to distinguish natural images (top left) and automatically generated image composites (bottom right). The red boundary illustrates the decision boundary between two. Our model is able to predict the {\em degree} of perceived visual realism of a photo, whether it's an actual natural photo, or a synthesized composite. For example, the composites close to the boundary appear more realistic. }
 \vspace{-5 mm}
\label{fig:teaser}
\end{figure}

In this paper, we are taking a small step in the direction of characterizing the space of natural images. We restrict the problem setting by choosing to ignore the issues of image layout, scene geometry, and semantics and focus purely on appearance.  For this, we use a large dataset of automatically generated image composites, which are created by swapping 
similarly-shaped object segments of the same object category between two natural images~\cite{lalonde2007using}.  This way, the semantics and scene layout of the resulting composites are kept constant, only the object appearance changes. Our goal is to predict whether a given image composite will be perceived as realistic by a human observer.
While this is admittedly a limited domain, we believe the problem still reveals the complexity and richness of our vast visual space, and therefore can give us insights about the structure of the manifold of natural images.

Our insight is to train a high-capacity discriminative model (a Convolutional Neural Network) to distinguish natural images (assumed to be realistic) from automatically-generated image composites (assumed to be unrealistic). 
Clearly, the latter assumption is not quite valid, as a small number of ``lucky'' composites will, in fact, appear as realistic as natural images. But this setup allows us to train on a very large visual dataset without the need of costly human labels.  One would reasonably worry that a classifier trained in this fashion might simply learn to distinguish natural images from composites, regardless of their perceived realism. But, interestingly, we have found that our model appears to be picking up on cues about visual realism, as demonstrated by its ability to rank image composites by their perceived realism, as measured by human subjects.  For example,  Figure~\ref{fig:teaser} shows two composites which our model placed close to the decision boundary -- these turn out to be composites which most of our human subjects thought were natural images. On the other hand, the composite far from the boundary is clearly seen by most as unrealistic.
Given a large corpus of natural and composite training images, we show that our trained model is able to predict the degree of realism of a new image. We observe that our model mainly characterizes the visual realism in terms of color, lighting and texture compatibility.

We also demonstrate that our learned model can be used as a tool for creating better image composites automatically via simple color adjustment.  Given a low-dimensional color mapping function, we directly optimize the visual realism score predicted by our CNN model. We show that this outperforms previous color adjustment methods on a large-scale human subjects study. We also demonstrate how our model can be used to choose an object from a category that best fits a given background at a specific location.
\section{Related Work}
\label{relatedwork}

Our work attempts to characterize properties of images that look realistic. This is closely related to the extensive literature on natural image statistics. Much of that work is based on generative models~\cite{Portilla03denoising,Elad2006sparsedict,Zoran2011gmmpatches}. Learning a generative model for full images is challenging due to their high dimensionality, so these works focus on modeling local properties via filter responses and small patch-based representations. These models work well for low-level imaging tasks such as denoising and deblurring, but they are inadequate for capturing higher level visual information required for assessing photo realism.

Other methods take a discriminative approach~\cite{ren2005data, HelOr2008wavelet,xue2012understanding,liu2013no,Schmidt2013discdeblur}. These methods can generally attain better results than generative ones by carefully simulating examples labeled with the parameters of the data generation process (e.g. joint velocity, blur kernel, noise level, color transformation).
Our approach is also discriminative, however, we generate the negative examples in a non-task-specific way and without recording the parameters of the process. Our intuition is that using large amounts of data leads to an emergent ability of the method to evaluate photo realism {\em from the data itself}.

In this work we demonstrate our method on the task of assessing realism of image composites. Traditional image compositing methods try to improve realism by suppressing artifacts that are specific to the compositing process. These include transition of colors from the foreground to the background~\cite{burt1983multiresolution,perez2003poisson}, color inconsistencies~\cite{Reinhard2001colortrans,reinhard2004real,lalonde2007using,xue2012understanding}, texture inconsistencies~\cite{johnson2011cg2real,Darabi2012melding}, and suppressing ``bleeding'' artifacts~\cite{tao2013error}. Some work best when the foreground mask aligns tightly with the contours of the foreground object~\cite{Reinhard2001colortrans,reinhard2004real,lalonde2007using,xue2012understanding}, while others need the foreground mask to be rather loose and the two backgrounds not too cluttered or too dissimilar~\cite{perez2003poisson,Hays2007,lalonde2007photo,tao2013error,Darabi2012melding}.  These methods show impressive visual results and some are used in popular image editing software like Adobe Photoshop, however they are based on hand-crafted heuristics and, more importantly, do not directly try to improve (or measure) the realism of their results. A recent work~\cite{Tan2015lighting} explored the perceptual realism of outdoor composites but focused only on lighting direction inconsistencies.

The work most related to ours, and a departure point for our approach, is Lalonde and Efros~\cite{lalonde2007using} who study color compatibility in image composites. They too generate a dataset of image composites and attempt to rank them on the basis of visual realism.  However, they use simple, hand-crafted color-histogram based features and do not do any learning.

Our method is also superficially related to work on digital image forensics~\cite{popescu2005exposing,kee2014exposing} that try to detect digital image manipulation operations such as image warping, cloning, and compositing, which are not perceptible to the human observer.
But, in fact, the goals of our work are entirely different: rather than detecting which of the realistic-looking images are fake, we want to predict which of the fake images will look realistic.
\section{Learning the Perception of Realism}
\label{sec:method}

Our goal is developing a model that could predict whether or not a given image will be judged to be realistic by a human observer.  However, training such a model directly would require a prohibitive amount of human-labeled data, since the negative (unrealistic) class is so vast.  Instead, our idea is to train a model for a different ``pretext'' task, which is: 1) similar to the original task, but 2) can be trained with large amounts of unsupervised (free) data.  The ``pretext'' task we propose is to discriminate between natural images and computer-generated image composites.
A high-capacity convolutional neural network (CNN) classifier is trained
using only automatically-generated ``free'' labels (i.e. natural vs. generated).
While this ``pretext'' task is different from the original task we wanted to solve (realistic vs. unrealistic), our experiments demonstrate that it performs surprisingly well on our manually-annotated test set (c.f. Section~\ref{sec:experiment}).

We use the network architecture of the recent VGG model~\cite{simonyan2014very}, a 16-layer model with small $3\times3$ convolution filters. We initialize the weights on the ImageNet classification challenge~\cite{deng2009imagenet} and then fine-tune on our binary classification task. We optimize the model using back-propagation with Stochastic Gradient Descent (SGD) using Caffe~\cite{jia2014caffe}.

\subsection{Automatically Generating Composites}
\label{subsec:data_generation}
\label{sec:data}

\begin{figure}[t]
\begin{center}
   \includegraphics[width=1.0\linewidth]{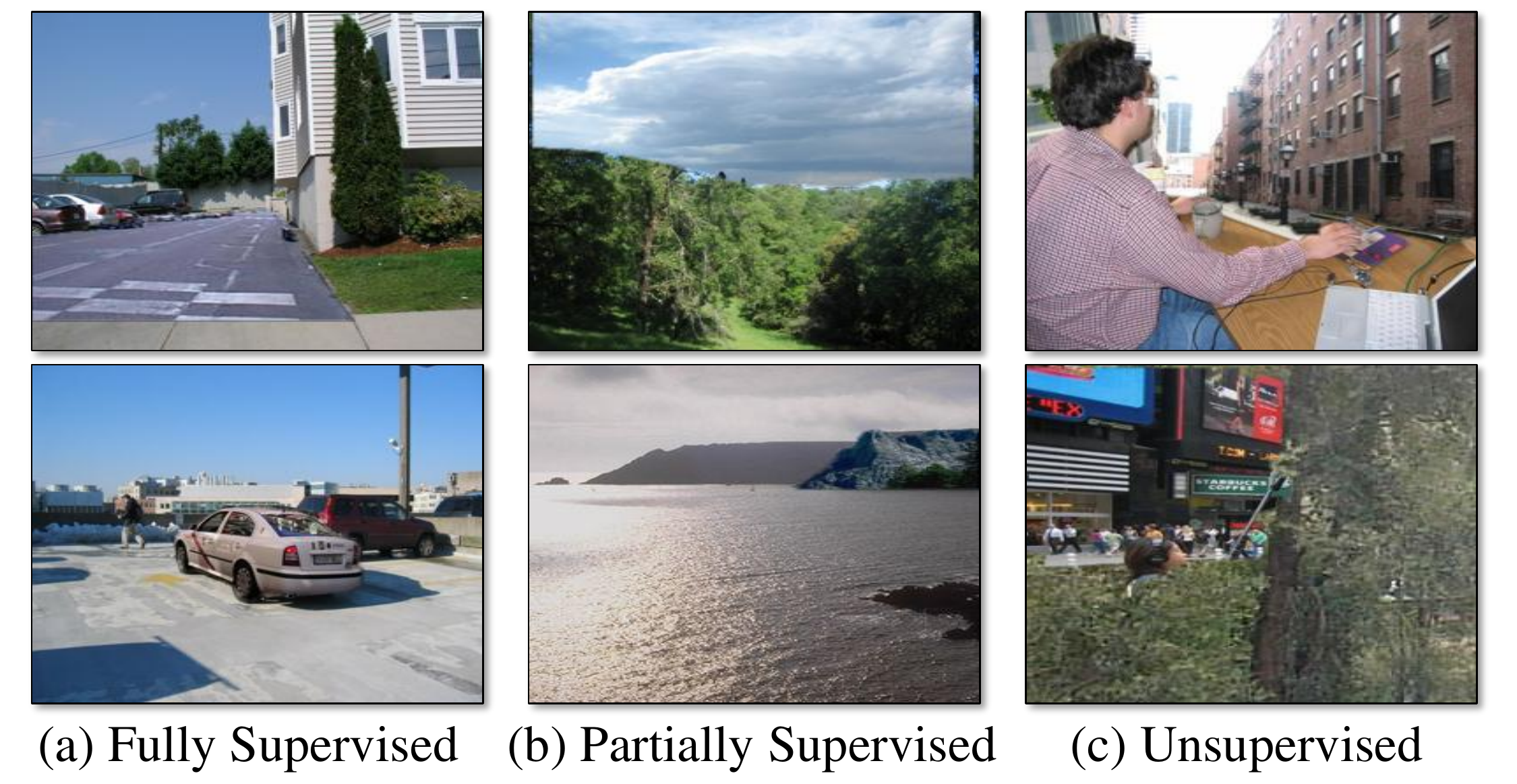}
\end{center}
 \vspace{-3 mm}
   \caption{Example composite images for CNN training: (a) image composites generated by fully supervised foreground and background masks, (b) image composites generated by a hybrid ground truth mask and object proposal, (c) image composites generated by a fully unsupervised proposal system. See text for details. Best viewed in color.}
\label{fig:data}
 \vspace{-3 mm}
\end{figure}
\begin{figure}
\begin{center}
 \includegraphics[width=1.0\linewidth]{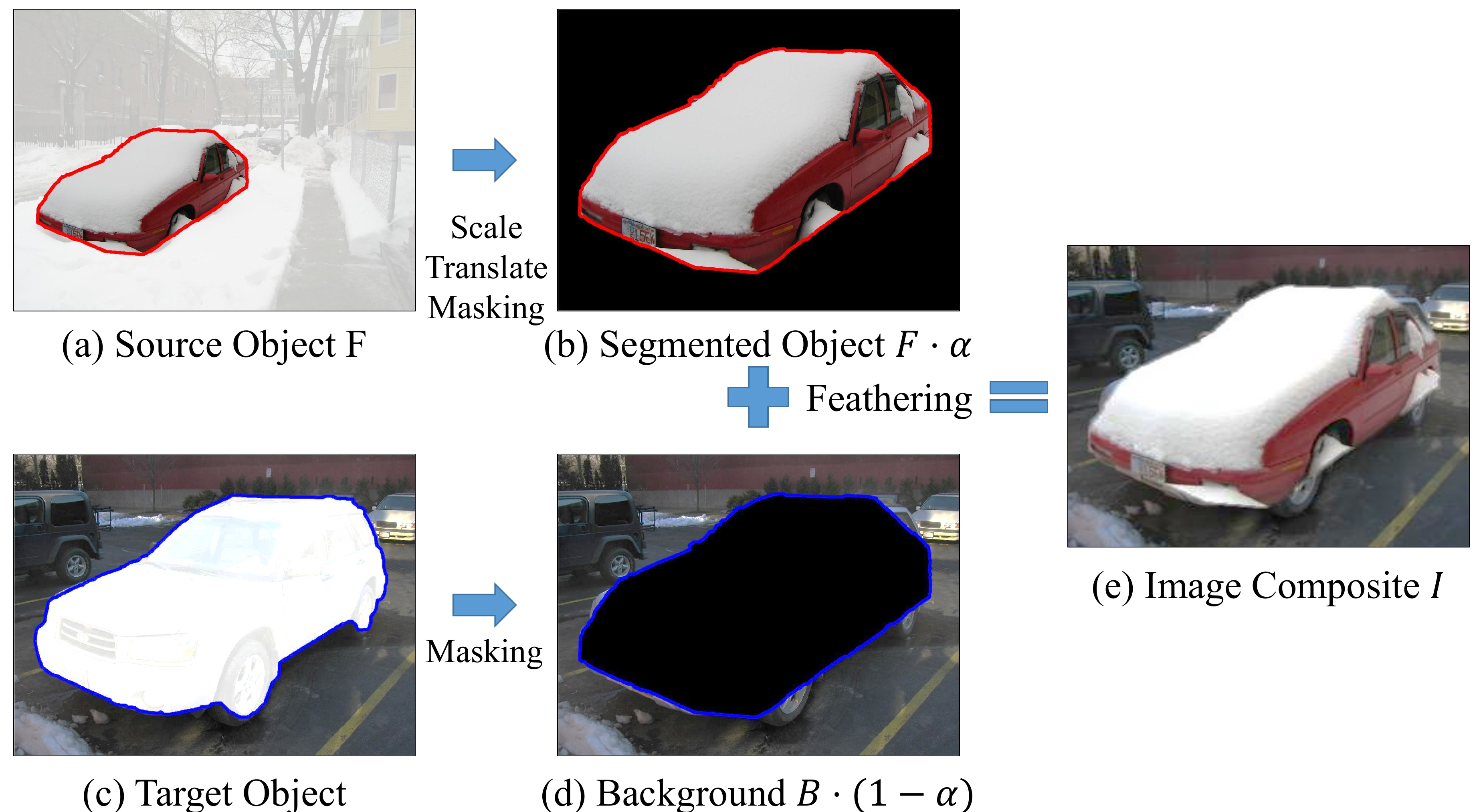}
\end{center}
 \vspace{-3 mm}
   \caption{We generate a composite image by replacing the target object (c) by the source object $F$ (a). We rescale and translate the source object to match the location and scale of the target object (c). We generate the final composite (e) by combining the segmented object (b) and the masked background (d).}
\label{fig:single}
 \vspace{-3 mm}
\end{figure}

To generate training data for the CNN model, we use the LabelMe image dataset~\cite{russell2008labelme} because it contains many categories along with detailed annotation for object segmentation. For each natural image in the LabelMe dataset, we generate a few composite images as follows.
 \vspace{-5 mm}
\paragraph{Generate a Single Composite}
Figure~\ref{fig:single} illustrates the process of generating a single composite image, which follows~\cite{lalonde2007using}. Starting with a background image $B$ (Figure~\ref{fig:single}c) that contains an object of interest (target object), we locate a source object $F$ (Figure~\ref{fig:single}a) with a similar shape elsewhere in the dataset, and then rescale and translate the source object $F$ so that the source object matches the target location. (Figure~\ref{fig:single}b). We assume the object is well segmented and the alpha map $\alpha$ of the source object is known (Figure~\ref{fig:single}d).
We apply a simple feathering based on a distance transform map to the object mask $\alpha$ of the source object. We generate the final composite by combining the source object and background $I = \alpha\cdot F+(1-\alpha)\cdot B$.

\begin{figure}
\begin{center}
  \includegraphics[width=1.0\linewidth]{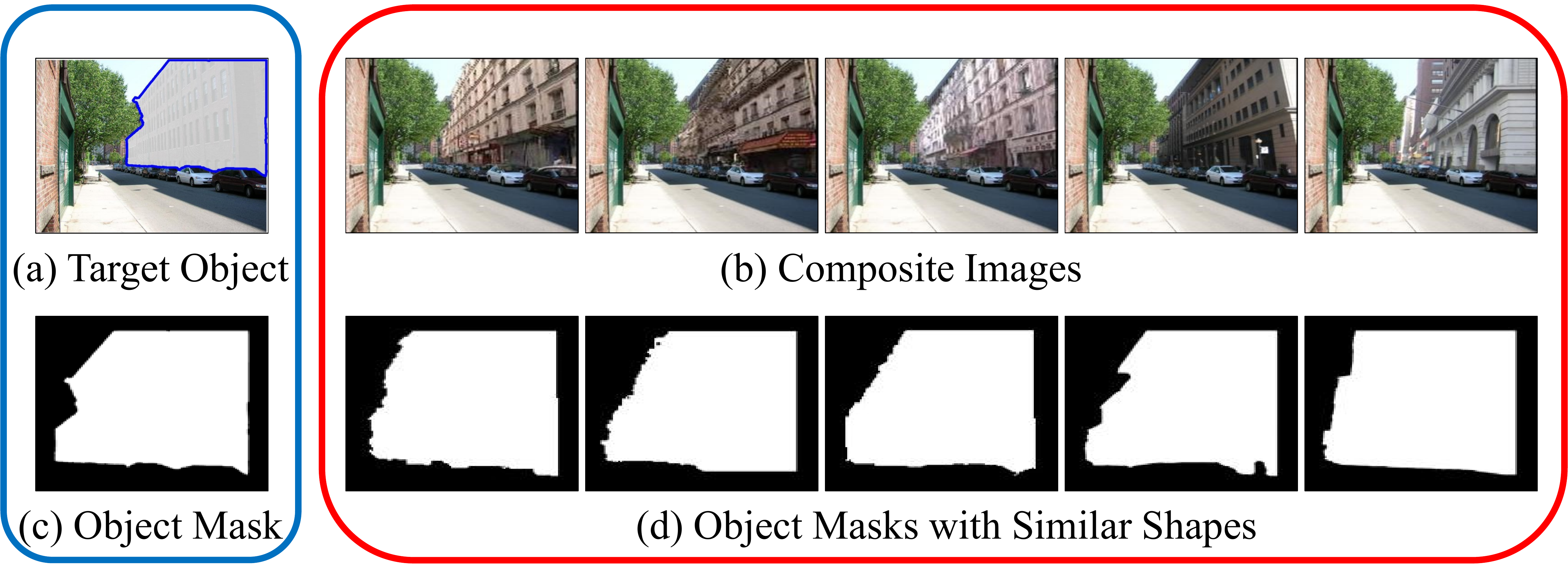}
\end{center}
 \vspace{-3 mm}
   \caption{Given an original photo with target object (a) and its object mask (c), we search for source objects whose object mask matches well the shape of target object, and replace the target object with them. We show the nearest neighbor object masks in (d) and their corresponding generated composites (b).}
\label{fig:data_generation}
 \vspace{-3 mm}
\end{figure}
 \vspace{-5 mm}
\paragraph{Generate Composite Dataset}
For each target object in each image, we search for source objects with similar shapes
by computing the SSD of blurred and subsampled ($64\times64$) object masks. Take Figure~\ref{fig:data_generation}, for example. We replace the original building with other buildings with similar outlines. The purpose of the rough matching of object shape is to make sure that the generated composites are already close to the manifold of natural images.
However, this procedure requires detailed segmentation annotations for both source and target objects. We call this procedure \textit{FullySupervised} as it requires full annotation of object masks.

An alternative way is to use automatic image segmentation produced by an ``object proposal'' method (in our implementation we used Geodesic Object Proposals~\cite{krahenbuhl2014geodesic}).
In this case, training images are still generated using human labeled segmentation for the target objects, but source objects are obtained by searching for object proposal segments with similar shapes to the target objects in all images.
This requires much fewer segmented training images.
We name this procedure \textit{PartiallySupervised}.
The third way is fully automatic: we use object proposals for both source and target objects. In particular, we randomly sample an object proposal for a given image, and replace it by other object proposals with the most similar shapes from the dataset.
This procedure is fully unsupervised and we call it \textit{Unsupervised}.
Later, we show that this fully automatic procedure only performs slightly worse than \textit{FullySupervised} w.r.t human annotations, in terms of predicting visual realism (Section~\ref{subsec:expr_classify}).
We also experimented with randomly cutting and pasting objects from one image to the other without matching object masks. In this case, the CNN model we trained mainly picked up artifacts of high-frequency edges that appear in image composites and performed significantly worse.
In our experiments, we used $\sim\!\!\!~11,000$ natural images containing $\sim\!\!\!~25,000$ object instances from the largest $15$ categories of objects in the LabelMe dataset. For \textit{FullySupervised} and \textit{PartiallySupervised}, we generated a composite image for each annotated object in the image. For \textit{Unsupervised}, we randomly sample a few object proposals as target objects, and generate a composite image for each of them.

Figure~\ref{fig:data} shows some examples of image composites generated by all three methods. Notice that some composite images are artifact-free and appear quite realistic, which forces the CNN model to pick up not only the artifacts of the segmentation and blending algorithms, but also the compatibility between the visual content of the inserted object and its surrounding scene. Different from previous work~\cite{lalonde2007using}, we do not manually remove any structurally inconsistent images. We find that composites generated by \textit{FullySupervised} are usually correct with regards to semantics and geometry, but sometimes suffer from inconsistent lighting and color.  \textit{PartiallySupervised} also often generates meaningful scenes, but sometimes tends to paste an object into parts of another object. While \textit{Unsupervised} tends to generate scenes with incorrect semantics, the number of scenes that can be generated is not restricted by the limited amount of human annotation.
\begin{figure}[t]
\begin{center}
    \begin{subfigure}[b]{1.0\linewidth}
                \includegraphics[width=\textwidth]{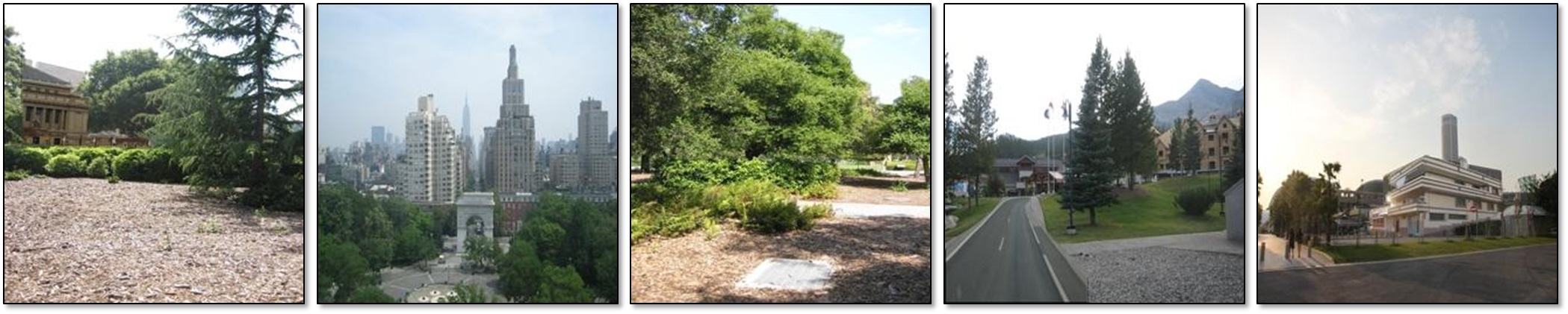}
                \caption{Most realistic composites ranked by our model}
                \label{fig:good_composite}
        \end{subfigure}  %
            \hfill
        \begin{subfigure}[b]{1.0\linewidth}
                \includegraphics[width=\textwidth]{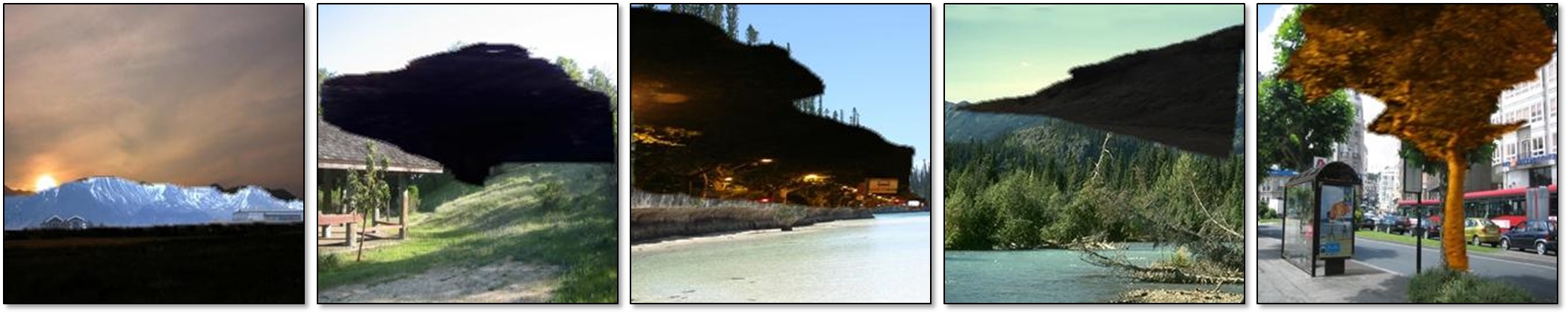}
                \caption{Least realistic composites ranked by our model}
                \label{fig:bad_composite}
        \end{subfigure}
\end{center}
 \vspace{-5 mm}
\caption{Ranking of generated composites in terms of realism scores. Best viewed in color.}
\label{fig:ranking}
 \vspace{-5 mm}
\end{figure}
 \vspace{-5 mm}
\paragraph{Ranking of Training Images}
Interestingly, our trained CNN model is able to rank visually appealing image composites higher than unrealistic photos with visual artifacts. In Figure~\ref{fig:ranking}, we use our model to rank the training composites by their realism score prediction. The top row shows high-quality composites that are difficult for humans to spot while the bottom row shows poor composites due to incorrect segmentation and color inconsistency.
We demonstrate that our model matches to human perception with quantitative experiments in Section~\ref{subsec:expr_classify}. 
\section{Improving Image Composites}
\label{sec:improve}

Let $f(I;\theta)$ be our trained CNN classifier model predicting the visual realism of an image $I$. We can use this classifier to guide an image compositing method to produce more realistic outputs.
This optimization not only improves object composition, but also reveals many of the properties of our learned realism model.

We formulate the object composition process as $I_g = \alpha \cdot g(F) + (1-\alpha) \cdot B$ where $F$ is the source object, $B$ is the background scene, and $\alpha \in [0,1]$ is the alpha mask for the foreground object.
For this task, we assume that the foreground object is well segmented and placed at a reasonable location.
The color adjustment model $g(\cdot)$ adjusts the visual properties of the foreground to be compatible with the background image.
Color plays an important role in the object composition process~\cite{lalonde2007using}. Even if an object fits well to the scene, the inconsistent lighting will destroy the illusion of realism.

The goal of a color adjustment is to optimize the adjustment model $g(\cdot)$, such that the resulting composite appears realistic.
We express this in the following objective function:
\begin{equation}
\label{eq:adjust}
E(g,F) = - f(I_g; \theta)+w \cdot E_{reg}(g),
 \end{equation}
where $f$ measures the visual realism of the composite and $E_{reg}$ imposes a regularizer on the space of possible adjustments.
A desired image composite should be realistic while staying true to identity of the original object (e.g. do not turn a white horse to be yellow).
The weight $w$ controls the relative importance between the two terms (we set it to $w=50$ in all our experiments).
We apply a very simple brightness and contrast model to the source object $F$ for each channel independently.
For each pixel we map the foreground color values $F^p=(c_1^p,c_2^p,c_3^p)$ to $g(F^p)=(\lambda_1c_1^p\!+\!\beta_1, \lambda_2c_2^p\!+\!\beta_2, \lambda_3c_3^p\!+\!\beta_3)$.
The regularization term for this model can be formulated as:
\begin{equation} \label{eq:reg}
\begin{split}
E_{reg}(g) &= \frac{1}{N}\sum_p \Big(\|I_g^p-I_0^p\|_2 +\\
&\sum_{i,j} \|(\lambda_i\!-\!1)\!\cdot\!c_i^p\!+\!\beta_i - (\lambda_j\!-\!1)\!\cdot\!c_j^p\!-\!\beta_j\|_2\Big)
\end{split}
\end{equation}

 \vspace{-2 mm}
where $N$ is the number of foreground pixels in the image, and $I_0=\alpha\cdot F + (1-\alpha)\cdot B$ is the composite image without recoloring, $I_g^p$ and $I_0^p$ denotes the color values for pixel p in the composite image. The first term penalizes large change between the original object and recolored object, and the second term discourages independent color channel variations (roughly  hue change).

Note that the discriminative model $\theta$ has been trained and fixed during this optimization.
 \vspace{-5 mm}
\paragraph{Optimizing Color Compatibility}
We would like to optimize color adjustment function $g^*=\arg \min_{g} E(g,F)$.
Our objective (Equation~\ref{eq:adjust}) is differentiable, if the color adjustment function $g$ is also differentiable.
This allows us to optimize for color adjustment using gradient-descent.

To optimize the function, we decompose the gradient into $\frac {\partial E}{\partial g} = -\frac{\partial f(I_g, \theta)}{\partial I_g} \cdot \frac{\partial I_g}{\partial g} +\frac{\partial E_{reg}}{\partial g}$. Notice that $ -\frac{\partial f(I_g,\theta)}{\partial I_g}$ can be computed through backpropagation of CNN model from the loss layer to the image layer while the other parts have a simple close form of gradient. See supplemental material for the gradient derivation.
We optimize the cost function using L-BFGS-B~\cite{byrd1995limited}.
Since the objective is non-convex, we start from multiple random initializations and output the solution with the minimal cost.

In Section~\ref{subsec:expr_optimize}, we compare our model to existing methods, and show that our method generates perceptually better composites. Although our color adjustment model is relatively simple, our learned CNN model provides guidance towards better color compatible composite.
 \vspace{-5 mm}
\paragraph{Selecting Best-fitting Objects}
Imagine that a user would like to place a car on a street scene (e.g. as in~\cite{lalonde2007photo}). Which car should she choose? We could choose an object $F^*=\arg \min_{F} E(g,F)$. For this, we essentially generate a composite image for each candidate car instance and select the object with minimum cost function (Equation~\ref{eq:adjust}). We show our model can select more suitable objects for composition task in Section~\ref{subsec:expr_select}.

\label{sec:optimize} 
\section{Implementation}
\label{sec:implementation}

\paragraph{CNN Training} We used the VGG model~\cite{simonyan2014very} from the authors' website, which is trained on ImageNet~\cite{deng2009imagenet}. We then fine-tune the VGG Net on our binary classification task (natural photos vs. composites). We optimize the CNN model using SGD. The learning rate $\alpha$ is initialized to $0.0001$ and reduced by factor 0.1 after $10,000$ iterations. We set the learning rate for $fc8$ layer to be $10$ times higher than the lower layers. The momentum is $0.9$, the batch size $50$, and the maximum number of iterations $25,000$.

 \vspace{-5 mm}
\paragraph{Dataset Generation}
For annotated objects and object proposals in the LabelMe dataset~\cite{russell2008labelme}, we only consider  objects whose pixels occupy between $5\%\sim 50\%$ of image pixels. For human annotation, we exclude occluded objects whose object label strings contain the words ``part'', ``occlude'', ``regions'' and ``crop''.
\section{Experiments}
\label{sec:experiment}

\label{subsec:expr_classify}

\begin{table}[t]
\begin{center}
 \begin{tabular}{l r}

  \toprule
  \multicolumn{2}{c}{Methods without object mask} \\
 \midrule
   Color Palette~\cite{lalonde2007using} (no mask) &  0.61 \\
 VGG Net~\cite{simonyan2014very} + SVM & 0.76 \\
 PlaceCNN~\cite{zhou2014learning} + SVM & 0.75\\
 AlexNet~\cite{krizhevsky2012imagenet} + SVM & 0.73 \\
 RealismCNN & 0.84 \\
 RealismCNN + SVM  & {\bf 0.88} \\
  Human & 0.91\\
  \midrule
  \multicolumn{2}{c}{Methods using object mask} \\
  \midrule
     Reinhard {\em et al.}~\cite{reinhard2004real} & 0.66\\
  Lalonde and Efros~\cite{lalonde2007using} (with mask) & 0.81 \\
  \bottomrule

\end{tabular}
 \vspace{-5 mm}
\end{center}
\caption {Area under ROC curve comparing our method against previous methods~\cite{reinhard2004real,lalonde2007using}.  Note that several methods take advantage of human annotation (object mask) as additional input while our method assumes no knowledge of the object mask.}
\label{tab:roc}
 \vspace{-5 mm}
\end{table}

\begin{figure*}
\begin{center}
 \includegraphics[width=0.9\linewidth]{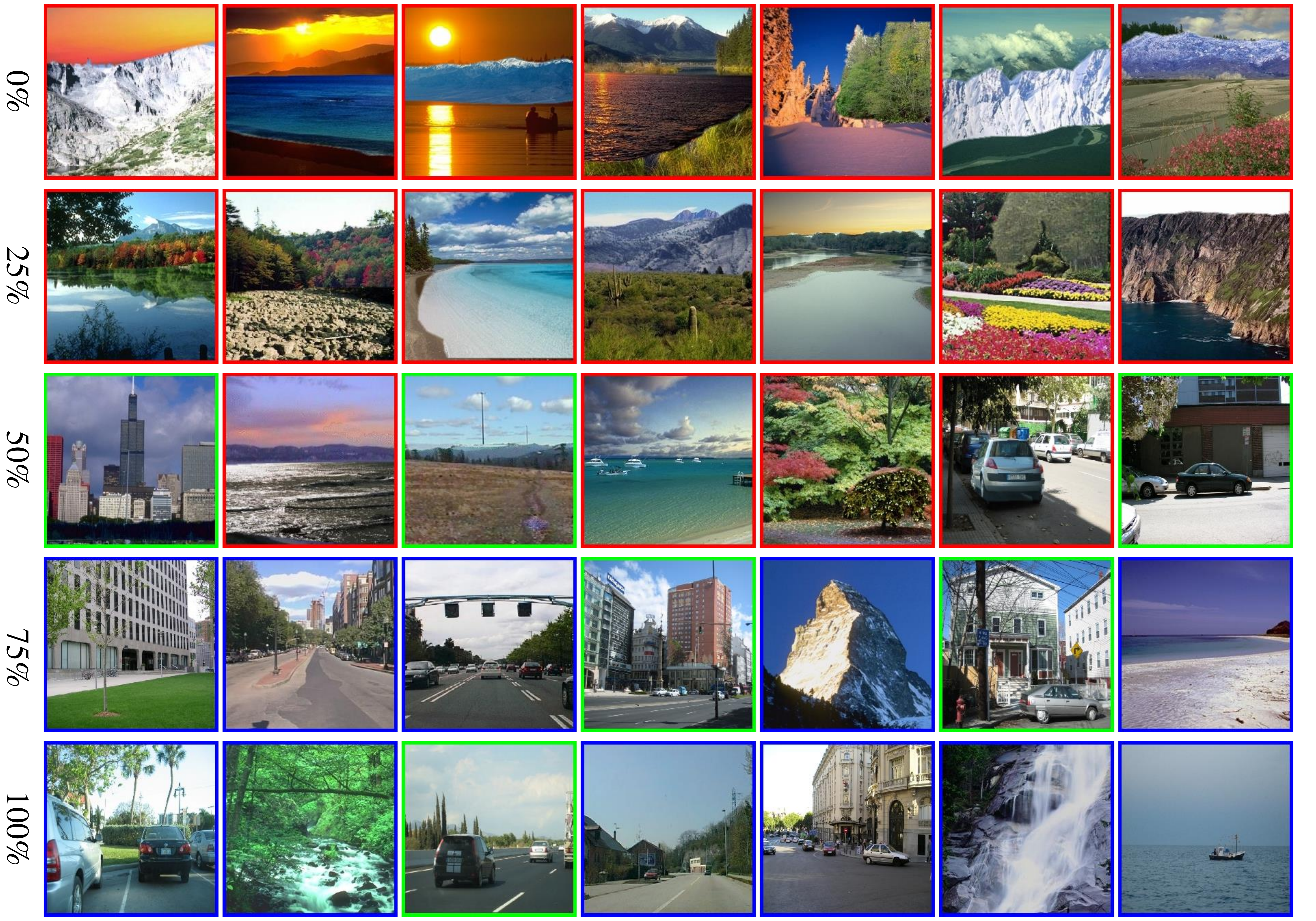}

 \vspace{-5 mm}
\end{center}
   \caption{Ranking of photos according to our model's visual realism prediction. The color of image border encodes the human annotation: {\color{green}green}: realistic composites; {\color{red}red}: unrealistic composites; {\color{blue}blue}: natural photos. The different rows contain composites corresponding to different rank percentiles of scores predicted with \textit{RealismCNN + SVM}.}
\label{fig:color_ranking}
 \vspace{-5 mm}
\end{figure*}

We first evaluate our trained CNN model in terms of classifying realistic photos vs. unrealistic ones.
 \vspace{-5 mm}
\paragraph{Evaluation Dataset} We use a public dataset of $719$ images introduced by Lalonde and Efros~\shortcite{lalonde2007using}, which comprises of $180$ natural photographs, $359$ unrealistic composites, and 180 realistic composites. The images were manually labeled by three human observers with normal color vision. All methods are evaluated on a binary realistic vs. unrealistic classification task with $359$ unrealistic photos versus $360$ realistic photos (which include natural images plus realistic composites).
Our method assigns a visual realism score to each photo. Area under ROC curve is used to evaluate the classification performance. We call our method \textit{RealismCNN}. Although trained on a different loss function (i.e. classifying natural photos vs. automatically generated image composites), with no human annotations for visual realism, our model outperforms previous methods that build on matching low-level visual statistics including color std/mean~\cite{reinhard2004real}, color palette, texture and color histogram~\shortcite{lalonde2007using}. Notice that Lalonde and Efros~\shortcite{lalonde2007using} also requires a mask for the inserted object, making the task much easier, but less useful.
 \vspace{-5 mm}
\paragraph{Supervised Training}
Without any human annotation for visual realism, our model already outperforms previous methods. But it would be more interesting to see how our \textit{RealismCNN} model improves with a small additional amount of human realism labeling. For this, we use the human annotation (realistic photos vs. unrealistic photos) provided by~\cite{lalonde2007using}, and train a linear SVM classifier~\cite{chang2011libsvm} on top of the $fc7$ layer's $4096$ dimensional features extracted by our \textit{RealismCNN} model, which is a common way to adapt a pre-trained deep model to a relatively small dataset. 
We call this \textit{RealismCNN + SVM}. Figure~\ref{fig:color_ranking} shows a few composites ranked with this model.
In practice, $fc6$ and $fc7$ layers give similar performance, and higher compared to lower layers. We evaluate our SVM model using 10-fold cross-validation. This adaptation further improves the accuracy of visual realism prediction. As shown in Table~\ref{tab:roc}, \textit{RealismCNN + SVM} ($0.88$) outperforms existing methods by a large margin. We also compare our SVM model with other SVM models trained on convolutional activation features ($fc7$ layer) extracted from different CNN models including AlexNet~\cite{krizhevsky2012imagenet} ($0.75$), PlaceCNN~\cite{zhou2014learning} ($0.73$) and original VGG Net~\cite{simonyan2014very} ($0.76$). As shown in Table~\ref{tab:roc}, our \textit{Realism + SVM} model reports much better results, which suggests that training a discriminative model using natural photos, and automatically generated image composites can help learn better feature representation for predicting visual realism.

 \vspace{-5 mm}
\paragraph{Human Performance} Judging an image as photo-realistic or not can be ambiguous even for humans. To measure the human performance on this task, we collected additional annotations for the $719$ images in~\cite{lalonde2007using} using Amazon Mechanical Turk. We collected on average $13$ annotations for each image by asking a simple question "Does this image look realistic?" and allowing the worker to choose one of four options: 1 (definitely unrealistic), 2 (probably unrealistic), 3 (probably realistic) and 4 (definitely realistic). We then average the scores of human response and compare the MT workers' ratings to the ``ground truth'' labels provided in the original dataset~\cite{lalonde2007using}.
Humans achieve a score of $0.91$ in terms of area under ROC curve, suggesting our model achieves performance that is close to level of human agreement on this dataset.

\begin{table}[t]
\begin{center}
 \begin{tabular}{l | c  c}
  & RealismCNN  & RealismCNN + SVM   \\
 \midrule
 \textit{FullySupervised}  & 0.84 & 0.88   \\
  \textit{PartiallySupervised} & 0.79 & 0.84  \\
 \textit{Unsupervised} & 0.78 & 0.84 \\

\end{tabular}

\end{center}
 \vspace{-5 mm}
\caption {Area under ROC curve comparing different dataset generation procedures. \textit{FullySupervised} uses annotated objects for both source object and target object. \textit{PartiallySupervised} uses annotated objects only for target object, but using object proposals for source object. \textit {Unsupervised}  uses object proposals for both cases.}
\label{tab:roc_generation}
 \vspace{-5 mm}
\end{table}

 \vspace{-5 mm}
\paragraph{Dataset Generation Procedure} The CNN we reported so far was trained on the image composites generated by the \textit{FullySupervised} procedure. In Table~\ref{tab:roc_generation}, we further compare the realism prediction performance when training with other procedures described in Section~\ref{subsec:data_generation}.
We find that \textit{FullySupervised} \textit{RealismCNN} gives better results when no human realism labeling is available. With SVM supervised training (using human annotations), the margin between different dataset generation methods becomes smaller. This suggests that we can learn the feature representation using fully unsupervised data (without any masks), and improve it using small amounts of human rating annotations.

 \vspace{-5 mm}
\paragraph{Indoor Scenes} The Lalonde and Efros dataset~\cite{lalonde2007using} contains mainly photographs of natural outdoor environments. To complement this dataset, we construct a new dataset that contains $720$ indoor photos with man-made objects from the LabelMe dataset. Similar to~\cite{lalonde2007using}, our new dataset contains $180$ natural photos, $180$ realistic composites, and $360$ unrealistic composites. To better model indoor scenes, we train our CNN model on  $\sim\!\!21,000$ natural images (both indoor and outdoor) that contain  $\sim\!\!42,000$ object instances from more than $200$ categories of objects in the LabelMe dataset. We use MTurk to collect human labels for realistic and unrealistic composites (13 annotations per image). Without SVM training, our $RealismCNN$ alone achieves $0.83$ on the indoor dataset, which is consistent with our results on the Lalonde and Efros dataset.

\subsection{Optimizing Color Compatibility }
\label{subsec:expr_optimize}

\begin{figure}
\begin{center}
 \includegraphics[width=1.0\linewidth]{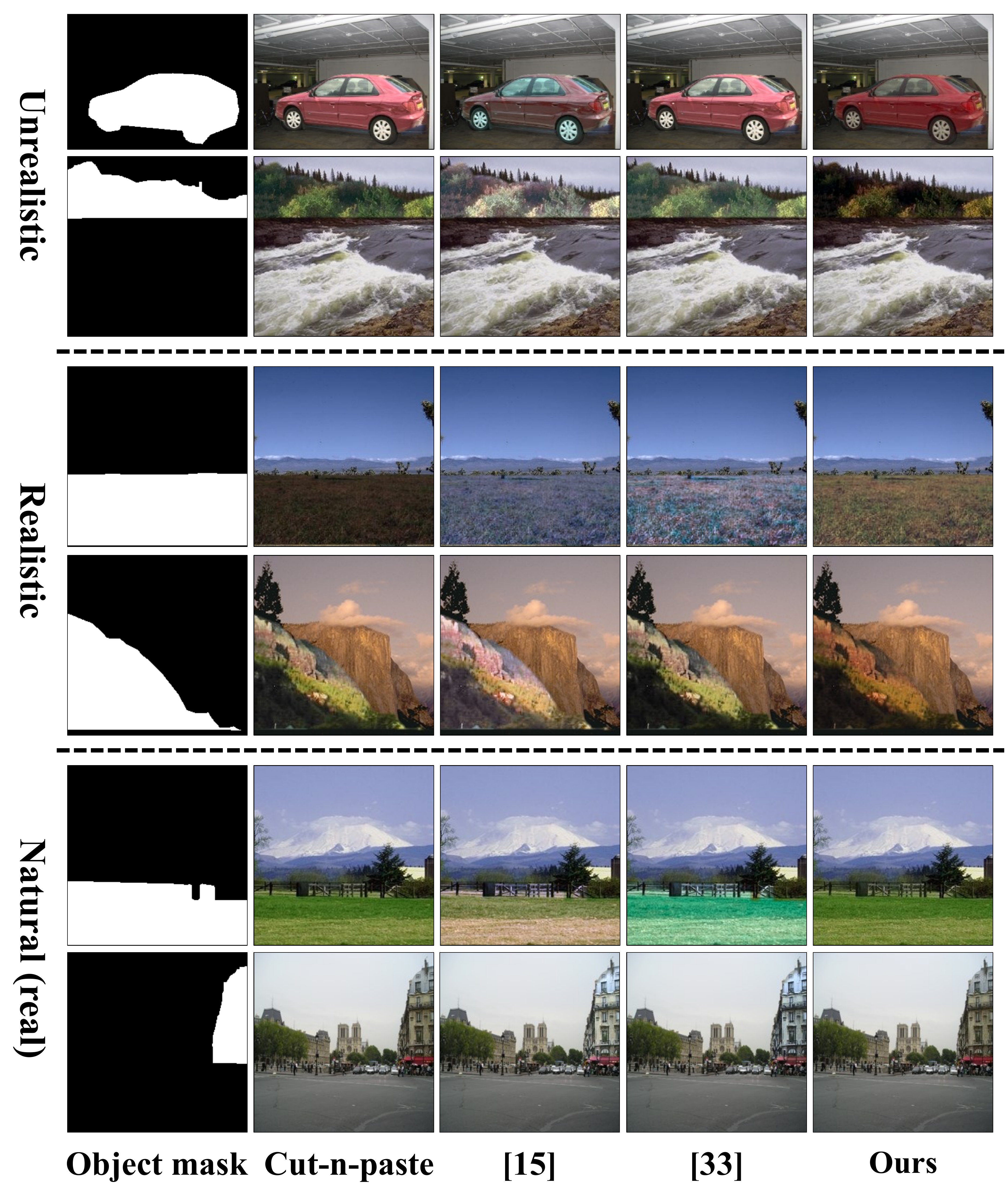}
\end{center}
 \vspace{-5 mm}
   \caption{Example composite results: from left to right: objects mask, cut-and-paste, Lalonde and Efros~\cite{lalonde2007using}, Xue et al.~\cite{xue2012understanding} and our method. }
\label{fig:color_adjust}
 \vspace{-5 mm}
\end{figure}

Generating a realistic composite is a challenging problem. Here we show how our model can recolor the object so that it better fits the background.
 \vspace{-5 mm}
 \paragraph{Dataset, Baselines and Evaluation} We use the dataset from~\cite{lalonde2007using} that provides a foreground object, its mask, and a background image for each photo. Given an input, we recolor the foreground object using four methods: simple cut-and-paste, Lalonde and Efros~\cite{lalonde2007using}, Xue et al.~\cite{xue2012understanding} and our color adjustment model described in Section~\ref{sec:improve}.  We use the \textit{FullySupervised} version of  \textit{RealismCNN} model without SVM training. We follow the same evaluation setting as in ~\cite{xue2012understanding} and use Amazon Mechanical Turk to collect pairwise comparisons between pairs of results (the question we ask is ``Given two photos generated by two different methods, which photo looks more realistic?''). We collected in total $43140$ pairwise annotations (10 annotations for each pair of methods for all $719$ images). We use the Thurstone's Case V Model~\cite{tsukida2011analyze} to obtain a realism score for each method per image from the pairwise annotations, and normalize the scores so that their standard deviation for each image is 1. Finally, we compute the average scores over all the photos. We report these average human rating scores for three categories of images: unrealistic composites, realistic composites and natural photos.  We use natural photos for sanity check since an ideal color adjustment algorithm should not modify the color distribution of an object in a natural photo. For natural photos, if no color adjustment is applied, the ``cut-and-paste'' result does not alter the original photo.

 \vspace{-5 mm}
\paragraph{Results}
Table~\ref{tab:color_adjust} compares different methods in terms of average human ratings. On average, our method outperforms other existing color adjustment methods. Our method significantly improves the visual realism of unrealistic photos. Interestingly, none of the methods can notably improve realistic composites although our model still performs best among the three color adjustment methods. Having a sense of visual realism informs our color adjustment model as to when, and how much, it should recolor the object. For both realistic composites and natural photos, our method typically does not change much the color distribution since these images are correctly predicted as already being quite realistic. On the other hand, the other two methods try to always match the low-level statistics between the foreground object and background, regardless of how realistic the photo is before recoloring.  Figure~\ref{fig:color_adjust} shows some example results.

\begin{figure}
\begin{center}
 \includegraphics[width=1.0\linewidth]{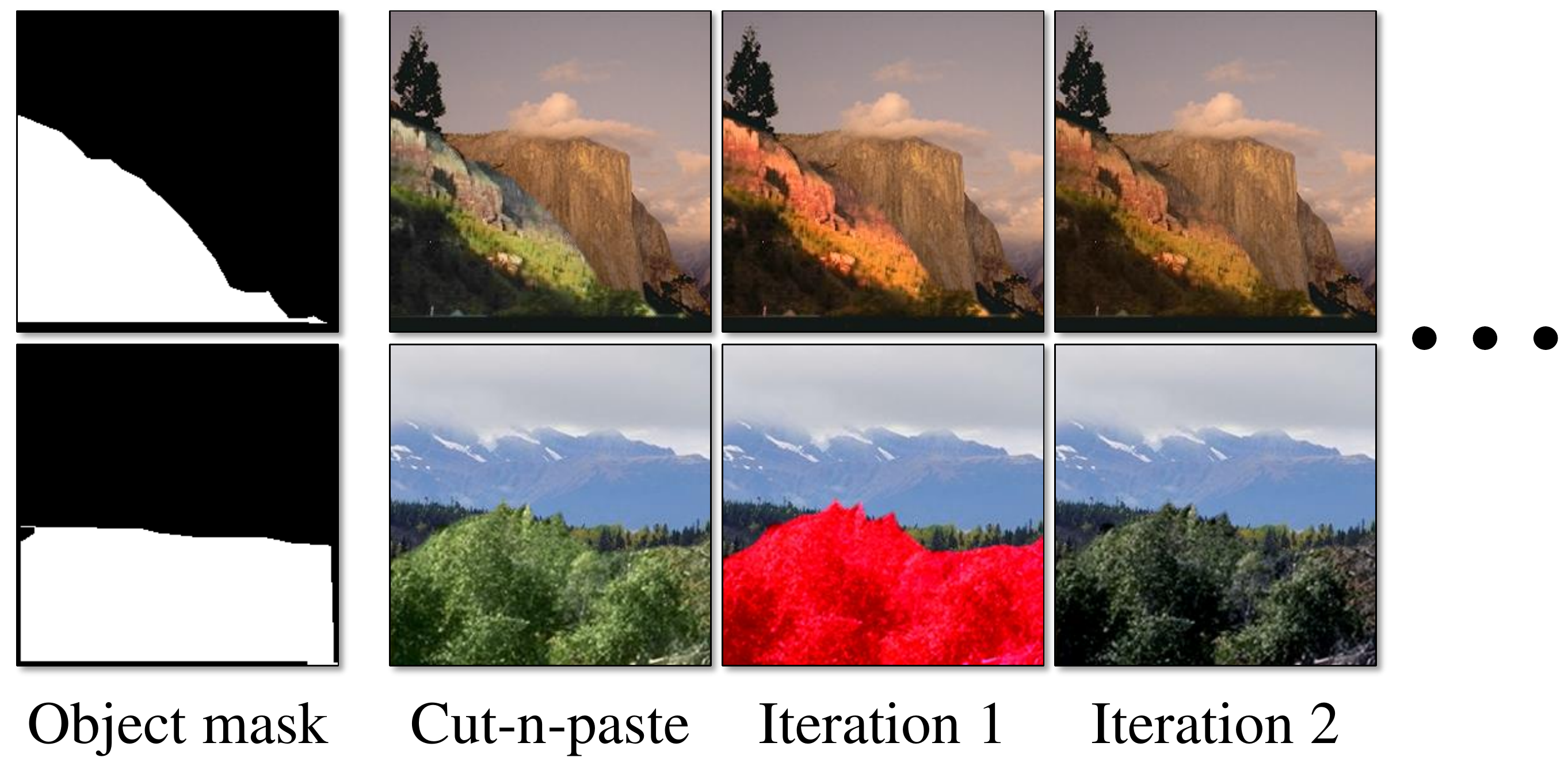}
\end{center}
 \vspace{-5 mm}
   \caption{From left to right: object mask, cut-and-paste, results generated by $CNNIter1$ and $CNNIter2$ without the regularization term $E_{reg}$.}
\label{fig:no_reg}
 \vspace{-5 mm}
\end{figure}

 \vspace{-5 mm}
\paragraph{Hard Negative Mining}
We observe that our color optimization method performs poorly for some images once we turn off the regularization term $E_{reg}$.
(See Figure~\ref{fig:no_reg} for examples). We think this is because some of the resulting colors (without $E_{reg}$) never appear in any training data (positive or negative). To avoid this unsatisfactory property, we add newly generated color adjustment results as the negative data, and retrain the CNN with newly added data, similar to hard negative mining in object detection literature~\cite{felzenszwalb2010object}. Then we use this new CNN model to recolor the object again. We repeat this process three times, and obtain three CNN models named as $CNNIter1$, $CNNIter2$ and $CNNIter3$.  We compare these three models (with $E_{reg}$ added back) using the same MTurk experiment setup, and obtain the following results: $CNNIter1$: $-0.162$, $CNNIter2$: $0.045$, and $CNNIter3$: $0.117$. As shown in Figure~\ref{fig:no_reg}, the hard negative mining avoids extreme coloring, and produces better results in general. We use $CNNIter3$ with $E_{reg}$ to produce the final results in Table~\ref{tab:color_adjust} and Figure~\ref{fig:color_adjust}.

\subsection{Selecting Suitable Object}
\label{subsec:expr_select}

We can also use our \textit{RealismCNN} model to select the best-fitting object from a database given a location and a background image. In particular, we generate multiple possible candidate composites for one category (e.g. a car) and use our model to select the most realistic one among them.

We randomly select $50$ images from each of the $15$ largest object categories in the LabelMe dataset and build a dataset of $750$ background images. For each background photo, we generate $25$ candidate composite images by finding $25$ source objects (from all other objects in the same category) with the most similar shapes to the target object, as described in Section~\ref{subsec:data_generation}. Then the task is to pick the object that fits the background best. We select the foreground object using three methods: using \textit{RealismCNN}, as described in Section~\ref{sec:improve}; select the object with the most similar shape (denoted \textit{Shape}); and randomly select the object from  $25$ candidates (denoted \textit{Random}).

\begin{figure}
 \centering
        \begin{subfigure}[b]{1.0\linewidth}
                \includegraphics[width=\textwidth]{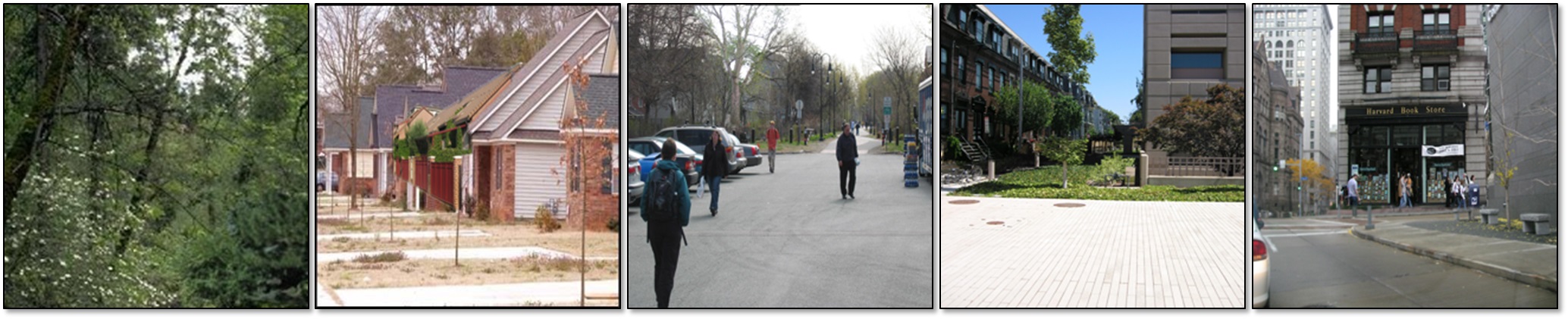}
 \vspace{-5 mm}
                \caption{Best-fitting object selected by \textit{RealismCNN}}
                \label{fig:object_rank_vgg}
        \end{subfigure}  %
            \hfill
        \begin{subfigure}[b]{1.0\linewidth}
                \includegraphics[width=\textwidth]{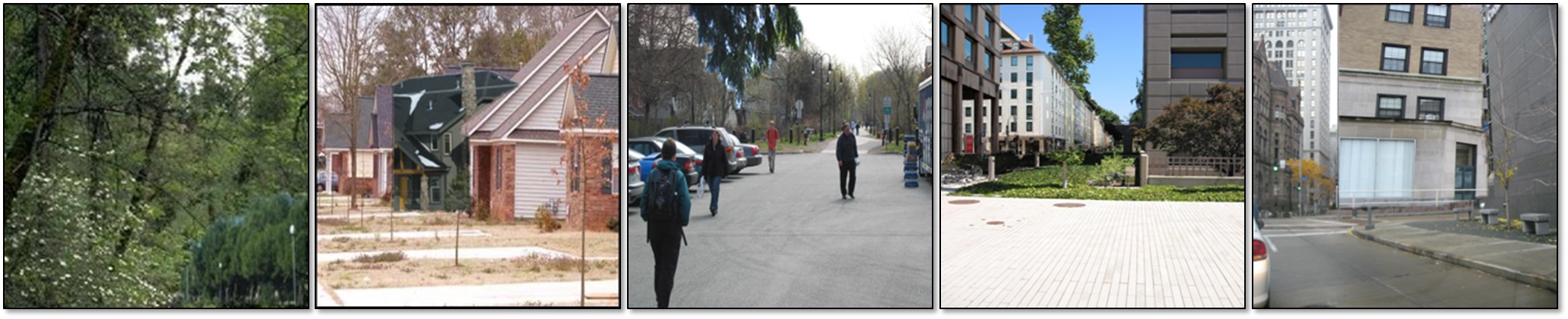}
 \vspace{-5 mm}
                \caption{Object with most similar shape}
                \label{fig:object_rank_shape}
        \end{subfigure}
         \begin{subfigure}[b]{1.0\linewidth}
                \includegraphics[width=\textwidth]{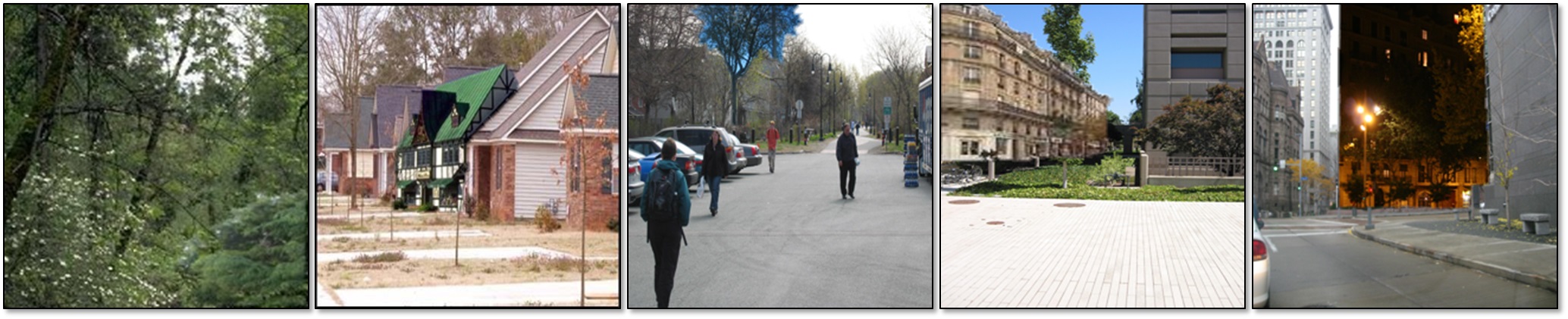}
 \vspace{-5 mm}
                \caption{Random selected objects}
                \label{fig:object_rank_rand}
        \end{subfigure}
 \vspace{-5 mm}
   \caption{For the same photo and the same location, we produce different composites using objects selected by three methods: (a) \textit{RealismCNN}, (b) the object with the most similar shape, and (c) a randomly selected object.}
\label{fig:object_rank}
\end{figure}

We follow the same evaluation setting described in Section~\ref{subsec:expr_optimize}.
We collect $22500$ human annotations, and obtain the following average Human ratings:
\textit{RealismCNN}: $0.285$, \textit{Shape}: $-0.033$, and \textit{Random}: $-0.252$. Figure~\ref{fig:object_rank} shows some example results for the different methods. Our method can suggest more suitable objects for the composition task.

\begin{table}[t]

\begin{center}
 \begin{tabular}{|c | c | c | c |}
 \hline
 & \begin{tabular}{@{}c@{}}Unrealistic \\ Composites\end{tabular} & \begin{tabular}{@{}c@{}}Realistic  \\ Composites\end{tabular} &  \begin{tabular}{@{}c@{}}Natural  \\ Photos\end{tabular} \\
 \hline
 cut-and-paste & -0.024& 0.263  &  0.287\\
 \hline
~\cite{lalonde2007using} &  0.123 & -0.299 & -0.247\\
 \hline
 ~\cite{xue2012understanding}  & -0.410 & -0.242&  -0.237 \\
 \hline
 ours & 0.311 & 0.279  & 0.196\\
  \hline

\end{tabular}

\end{center}
 \vspace{-5 mm}
\caption {Comparison of methods for improving composites by average human ratings. We use the authors' code to produce results for Lalonde and Efros~\cite{lalonde2007using} and Xue et al~\cite{xue2012understanding}. We follow the same evaluation setting as in~\cite{xue2012understanding} and obtain human ratings from pairwise comparisons using Thurstone's Case V Model~\cite{tsukida2011analyze}.}
\label{tab:color_adjust}
 \vspace{-5 mm}
\end{table} 
\section{Conclusion}
\label{sec:conclusion}
In this paper, we present a learning approach for characterizing the space of natural images, using a large dataset of automatically created image composites. We show that our learned model can predict whether a given image composite will be perceived as realistic or not by a human observer. Our model can also guide automatic color adjustment and object selection for image compositing.

Many factors play a role in the perception of realism. While our learned model mainly picks up on purely visual cues such as color compatibility, lighting consistency, and segment compatibility, high-level scene cues (semantics, scene layout, perspective) are also important factors. Our current model is not capable of capturing these cues as we generate composites by replacing the object with an object from the same category and with a similar shape. Further investigation in these high-level cues will be required.

 \vspace{-5 mm}
\paragraph{Acknowledgements}
We thank Jean-Fran\c{c}ois Lalonde and Xue Su for help with running their code. This work was sponsored in part by ONR MURI N000141010934, an Adobe research grant, a NVIDIA hardware grant and an Intel research grant. J.-Y. Zhu was supported by Facebook Graduate Fellowship.

{\small
\bibliographystyle{ieee}
\bibliography{tex}
}

\end{document}